\documentclass{article}
\usepackage{ijcai}

\usepackage{times}
\usepackage{soul}
\usepackage{url}
\usepackage[hidelinks]{hyperref}
\usepackage[utf8]{inputenc}
\usepackage[small]{caption}
\usepackage{graphicx}
\usepackage{amsmath}
\usepackage{amsthm}
\usepackage{booktabs}
\usepackage{algorithm}
\usepackage{algorithmic}
\usepackage{subfig}
\usepackage{multirow}
\usepackage{amsfonts,amssymb}
\usepackage{color}

\title{DepthGAN: GAN-based Depth Generation of Indoor Scenes from Semantic Layouts}

\author{
Yidi Li$^1$
\and
Yiqun Wang$^2$\and
Zhengda Lu$^{1}$\And
Jun Xiao$^1$
\affiliations
$^1$School of Artificial Intelligence, University of Chinese Academy of Sciences\\
$^2$School of Computer Science, Chongqing University\\
\emails
liyidi19@mails.ucas.ac.cn,
csyqwang@hotmail.com,
\{luzhengda, xiaojun\}@ucas.ac.cn
}

\begin{document}

\maketitle

\begin{abstract}


Limited by the computational efficiency and accuracy, generating complex 3D scenes remains a challenging problem for existing generation networks. 
In this work, we propose DepthGAN, a novel method of generating depth maps with only semantic layouts as input. First, we introduce a well-designed cascade of transformer blocks as our generator to capture the structural correlations in depth maps, which makes a balance between global feature aggregation and local attention.  
Meanwhile, we propose a cross-attention fusion module to guide edge preservation efficiently in depth generation, which exploits additional appearance supervision information. 
Finally, we conduct extensive experiments on the perspective views of the Structured3d panorama dataset and demonstrate that our DepthGAN achieves superior performance both on quantitative results and visual effects in the depth generation task.
Furthermore, 3D indoor scenes can be reconstructed by our generated depth maps with reasonable structure and spatial coherency.

\end{abstract}

\section{Introduction}


3D scene generation is an important task of computer vision, which can be used in a variety of downstream applications, such as 3D scene modeling, AR and VR, etc.
However, existing 3D generation methods mainly focus on constructing single objects \cite{ref:diff-point-gen,ref:GenPointNet} or optimizing the layout of existing 3D models for scene construction \cite{ref:3D-sln} due to the limitation of computational resources and the complexity of object relations in 3D scenes. 
Thus, generating complex 3D indoor scenes containing different objects and stuff remains a challenging problem.


Rather than manually building 3D scenes with multiple objects, visual designers typically prefer controlled and simple input, such as sketches or layouts. However, due to insufficient input information, it is difficult to directly construct 3D scenes. Inspired by 3D reconstruction tasks, the depth map is a viable medium, which can be regarded as the transition from 2D layouts to 3D scenes.
In this work, we aim to generate 3D scenes in the form of depth maps from the layout inputs with efficient semantic and relational constraints. Since the depth map provides accurate geometric relations within the scene, 
3D scenes can be further constructed.
\begin{figure}[tbp]
    \begin{center}
	    \includegraphics[width=0.99\linewidth]{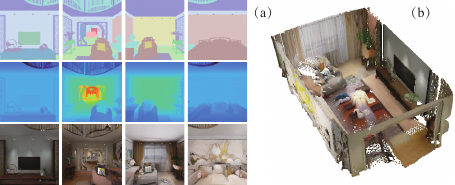}
    \end{center}
    \caption{The construction results of our model. (a) input semantic layouts with our generated depth maps and RGB images. (b) point clouds reconstructed from our generated depth maps and RGB images. More details can be found in the supplementary material.}
	\label{fig:pointcloud}
\end{figure}





For this purpose, we first test the depth generation on mainstream conditional image generation models \cite{ref:pix2pixHD,ref:SPADE} based on CNNs. However, convolution operations are proven to have the same effect as filters with learnable parameters, and CNN-based generation models significantly eliminate the components with high frequency as demonstrated in \cite{ref:frequency_loss}. 
Therefore, the multi-layer convolution architecture filters out the boundaries of different objects and leads to the unclear hierarchical structure and boundaries in the generated depth maps.


Meanwhile, the receptive field of CNNs is limited in the local scope \cite{ref:receptive-field}.
Hence most existing CNN-based methods for semantic image generation can not accurately predict global correlations even with pre-designed optimizations on global features.
And the attention is paid to pixels in the receptive field rather than those outside the sight, which makes the generated results incoherent in terms of visual perception.
Moreover, the depth map emphasizes the structural relationship between different objects, and this defect causes these methods to generate incorrect structural correlations. 


In this paper, we propose the DepthGAN to efficiently generate depth maps for indoor scenes from the input of semantic layouts. 
First, we utilize a well-designed cascade of transformer blocks with conditional normalization on tokens in our generator to capture the structural correlation of depth maps. Especially, we focus more on the global feature generation while along with mimicking the receptive field of the convolutional architecture for local features.
Meanwhile, depth maps emphasize the smoothness in the interior of the object more than color images. We use a learned residual connection with 2D convolutional normalization to smooth the depth generation of inside objects. 
Furthermore, we propose a cross attention fusion (CAF) module to preserve the semantic boundaries of depth maps, which facilitates the consistency of depth structure with the supervision of RGB images.
Finally, we can also reconstruct the 3D indoor scenes by our generated depth maps with the reasonable structure and spatial coherency as shown in Fig.\ref{fig:pointcloud}.
The contributions of this paper are summarized as follows:

\begin{itemize}
    \item[1.] We introduce a novel approach of generating 3D scenes with semantic layout inputs using depth maps as intermediate guidance. It provides an effective solution for 3D structures generation of complex scenes.
    \item[2.] We propose a cascaded transformer-based generator for the pipeline from semantic layouts to depth generation. Our model generates depth maps with global structures and reasonable structural correlation, both in terms of quantitative metrics and visual effects. 
    \item[3.] A cross attention fusion module is proposed to exploit the supervision information of color images. The fusion guidance using color images can effectively preserve semantic edges in the depth generation results.
 \end{itemize}
\section{Related Work}
\paragraph{Depth Generation.}
Existing depth generation methods mainly focus on regressing dense depth maps from images such as monocular depth estimation \cite{ref:AdaBins,ref:BTS}. 
Recent works have proposed to exploit assumptions about indoor scenes, such as planarity constraints\cite{ref:BTS}, to guide the network in an encoder-decoder architecture. 
Nevertheless, poor edge quality and the lack of global information are common problems of CNN-based depth estimation models. \cite{ref:edge-depth} explicitly uses pre-trained semantic segmentation to guide depth boundaries due to the high quality of edges in the semantic map. 
\cite{ref:AdaBins} performs a global statistical analysis of the output of encoder-decoder architecture for refining the output depth at the highest resolution.
The regressed depth map can be further used to reconstruct the 3D scene, which inspires us a time-saving choice to generate 3D scenes by means of depth maps. 
However, depth estimation tasks typically require enough natural images as input, which is hard to meet the visual designers' requirements of the simplicity of manipulation and variability.
To solve this problem, we utilize generative models with the input of semantic layouts for the depth generation task.

\paragraph{Generative Models.}
Generative models have been applied widely in various tasks such as conditional image generation \cite{ref:pix2pixHD,ref:SPADE} and 3D model generation \cite{ref:GenPointNet,ref:SP-GAN}. 
Conditional image generation aims to generate photorealistic images using conditional input such as texts and semantic layouts, which has been boosted to high resolution.
However, the depth maps generated by pix2pixHD and SPADE tend to be over-smoothed, resulting in ambiguous depth boundaries. Meanwhile, the local receptive fields of CNN lead to incorrect depth relationships.
\cite{ref:CC-FPSE} predict depthwise separable convolution conditioned on the semantic layouts to keep semantic consistence the generated results, which also helps for edge information retention.
On the other hand, existing 3D generation models mainly focus on the small scale objects such as \cite{ref:usl-3d-gen-by-2d} generates 3D objects by 2D projection matching.
But this is difficult to directly apply to large-scale 3D scenes generation due to the limitation of model resolution.

Therefore, we utilize the vision transformer for depth generation to make a balance between global feature aggregation and local attention and preserve the global depth structures.

\paragraph{Vision Transformer}
Vision Transformer(ViT) \cite{ref:vit} is a self-attention model using image patches for local feature aggregation and extraction, which is helpful to capture long-range dependencies in images and recover global structure.
\cite{ref:transgan,ref:vitgan,ref:improved_tran_gan} proposed GAN models based on ViT for unconditional image generation. 
However, the generation quality of these methods is not proportional to the time consumption due to the low computing efficiency of ViT. 
Swin Transformer \cite{ref:swin-transformer} improves the computational efficiency and accuracy of vision transformer by the shifted window multi-head self-attention strategy, which makes it possible to apply vision transformers for generation tasks. 
\cite{ref:styleswin} proves the promise of using swin transformer for high-resolution image generation. 

In this work, we focus on conditional generation tasks for converting the input semantic layouts to a depth map which contains precise structural information. 
To the best of our knowledge, this is the first work that explores depth generation frameworks for 3D scenes that only use layouts as input.


\section{Method}
\begin{figure*}[htbp]
    \begin{center}
	    \includegraphics[width=0.85\linewidth]{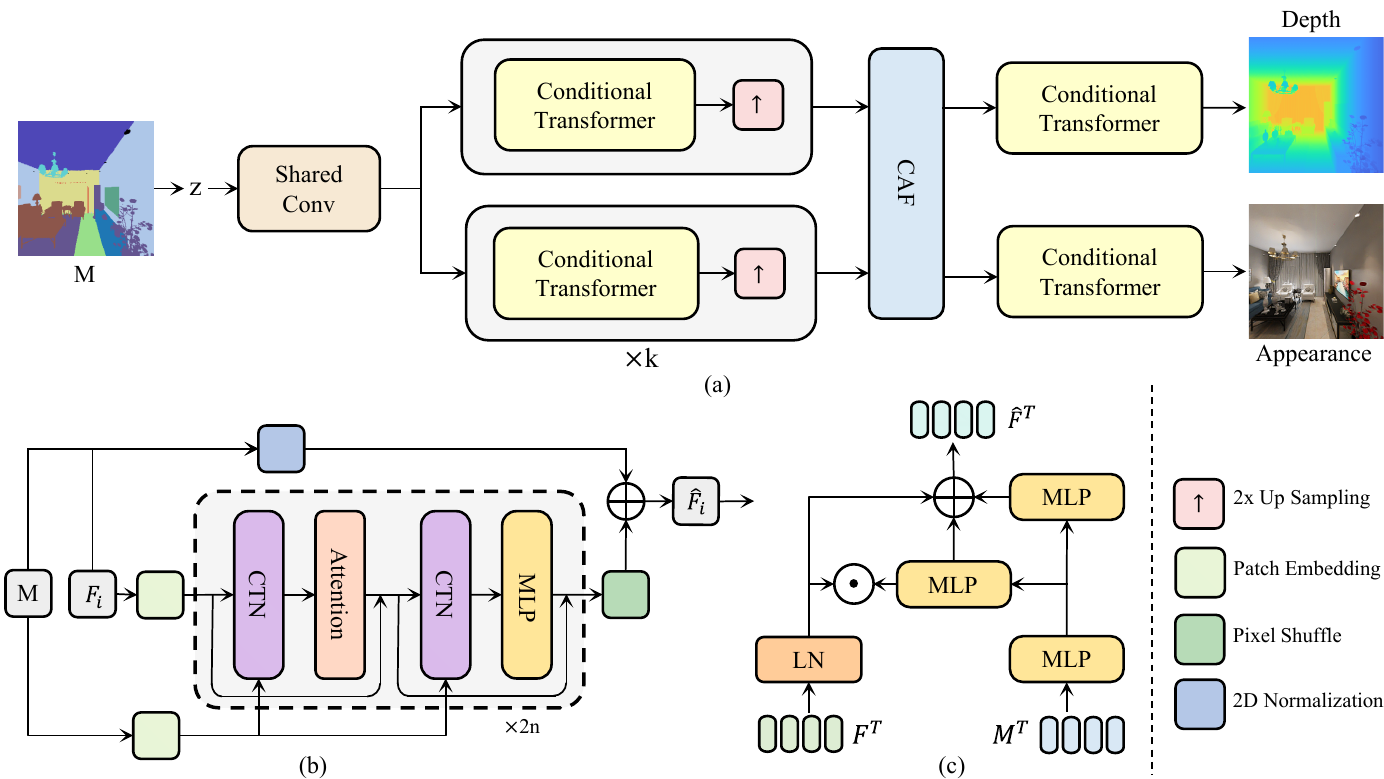}
    \end{center}
    \caption{Overview of the generator of our proposed DepthGAN. (a) pipeline of generator; (b) swin transformer based generator block; (c) conditional token normalization module.}
	\label{fig:pipeline}
\end{figure*}

Let $M \in \mathbb{L}^{H\times W}$ be a semantic layout where $\mathbb{L}$ is a set of integers denoting the semantic label of each pixel, and $H,W$ are the height and width of the semantic layout. 
As illustrated in Fig.\ref{fig:pipeline} (a), the generator of our proposed DepthGAN, which takes the $M$ as input and gets the depth map $D \in \mathbb{R}^{H \times W}$ as output.
First, we encode the input $M$ into a latent vector \text{z} and utilize a shared convolutional module to generate the conjoint features of depth and RGB images.
After that, we use the cascaded swin transformer-based generator blocks with up-sampling to enable the coarse-to-fine generation.
Especially, we use two branches for generating depth maps and RGB images simultaneously and we fuse the features of two branches with the CAF module as the cross-domain guidance after the penultimate stage.

\subsection{Swin Transformer based Generator Block.}
Generally, traditional CNN-based methods cannot effectively interact with the overall depth structure.
Therefore, we adopt the swin transformer \cite{ref:swin-transformer} as the basic generator block compared with the ViT\cite{ref:vit} with higher computational complexity.
Our swin transformer based generator block takes the semantic layout $M$ and feature map $F_i$ from the previous stage in different scales as the inputs at each stage as shown in Fig.\ref{fig:pipeline} (b).



Given an feature map $F_i \in \mathbb{R}^{C_i \times H_i\times W_i}$ at the $i$-th stage, where $C_i$ is the number of channels and $H_i$ and $W_i$ are width and height respectively. 
We first utilize a series of swin transformer blocks to generate the details of tokens with multi-head self attention $\text{MSA}$ \cite{ref:swin-transformer} which uses the no-shifted or shifted window partition strategy to advocate the information interaction among consecutive windows. Meanwhile, we replace the layer normalization module in the swin transformer with our Conditional Token Normalization(CTN) module which fuses the semantic constraints to the tokenized feature maps. 
The transformer architecture can be described as:
\begin{equation}
    F_{i,l+1}^{T} = \text{MLP}(\text{CTN}(\text{Attn}(\text{CTN}(F_{i,l}^{T}, M)),M))\\
\end{equation}
where $F_{i,l}^{T}$ is the input tokens of the $l$-th layer at the $i$-th stage.  \text{Attn} denotes the attention blocks with the window-based multi-head self-attention under regular (\text{W-MSA}) or shifted window partition (\text{SW-MSA}), and always appears in pairs \cite{ref:swin-transformer}. Note that the number of \text{W-MSA} and \text{SW-MSA} pairs is set to be even.

After the consecutive attention operations, we reshape the sequence of tokens $F^T_i$ back to a 2D feature map and then adopt a \text{pixelshuffle} module \cite{ref:pixelshuffle} to upsample its resolution and downsample the embedding dimension if the feature map is patch embedded. The output 2D feature map is calculated as:
\begin{equation}
    \hat{F}_i = \text{PixelShuffle}(\text{Reshape}(F^{T}_i))
\end{equation}


Then, we learn a residual connection between $F_i$ and $\hat{F}_i$. Due to the inconsistency of channels, we simply apply the learned SPADE normalization \cite{ref:SPADE}
for $F_i$ using the input semantic layout $M$. The learned shortcut matches the channels of $F_i$ to $\hat{F}_i$ and provides more 2D information apart from 1D tokens, which also helps smooth the depth map. 

Finally, the feature map is upsampled and fed to the next stage.
Specially, we gradually increase the patch size and window size at each stage along with the increasing resolution of feature maps. Thus, our well-designed cascaded swin transformer blocks enable the model to accurately generate structure features and achieve a similar receptive field of CNNs. 
Moreover, we use more cascade pairs of swin blocks in higher input resolution, which is beneficial for generating more natural features.
Therefore, in our generator module, we achieve a trade-off between computational complexity with generating capacity and the local attention with global feature aggregation.


\subsection{Conditional Token Normalization.}  
Due to existing ViT based generation models are unconditional, we introduce the Conditional Token Normalization(CTN) method to learn the affine transformation with the input condition in the layer normalization.


With the input tokens $F^T$ with length $L$ and embedding dim $E$, our CTN module is formulated as:
\begin{equation}
\hat{F}^T = \frac{\gamma(M^{T})}{\sigma}\odot (F^{T}-\mu)+\beta(M^{T})
\end{equation}
where $\odot$  is the element-wise multiplication between two vectors. $\mu$ and $\sigma$ denote the mean and standard deviation of $F^{T}$:
\begin{align}
& \mu = \frac{1}{EL} \sum \limits_{C,L} F^{T}\\
& \sigma = \sqrt{\frac{1}{EL} \sum \limits_{C,L} (F^{T}-\mu)^{2}}
\end{align}

As shown in Fig.\ref{fig:pipeline} (c), we apply the layer norm to $F^{T}$ without the affine transformation and use a MLP to learn the hidden representation of tokenized semantic layout $M^{T}$. After that, the vector $\gamma(M^{T})$ and $\beta(M^{T})$ are learned from the hidden representation using MLPs, which aims to formulate a conditional layer normalization in tokenized feature maps. 
Finally, CTN make effective use of the input condition in the ViT for our generator.


\subsection{Cross Attention Fusion module}
\begin{figure}[htbp]
    \begin{center}
	    \includegraphics[width=0.35\textwidth]{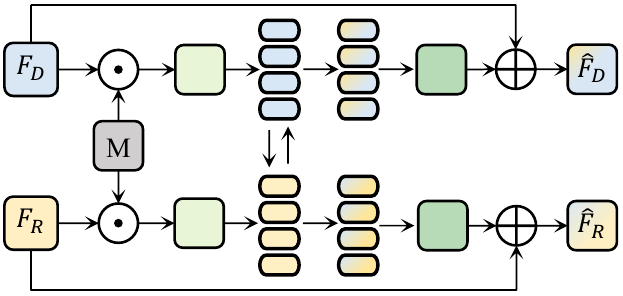}
    \end{center}
    \caption{Cross attention fusion module. $\odot$ is element-wise multiplication.}
	\label{fig:caf}
\end{figure}

Due to the consistency of some properties between RGB and depth domains, such as the semantic boundaries, we employ a CAF module to enhance the consistent features by the supervision of RGB images, as shown in Fig.\ref{fig:caf}. 


First, we downsample and embed $M$ to the same spatial resolution and dimension as the current feature map. 
Then we perform a semantic alignment by the element-wise multiplication between two input feature maps and the embedded semantic layout.
Thus, we fuse the semantic information for the further cross attention.
For computational efficiency, we adopt patch embedding for the aligned feature map. $F_D$ and $F_R$ denote the tokenized feature maps in the depth and RGB branches, respectively. 
Our cross-attention module is formulated as:
\begin{align}
& \hat{F}_D = \text{Softmax}(\frac{Q_DK_R}{\sqrt{d}})V_D \\
& \hat{F}_R = \text{Softmax}(\frac{Q_RK_D}{\sqrt{d}})V_R
\end{align}
where $Q,K,V,d$ are projected query, key, value matrices and dimension in attention operation. $\hat{F}_D$ and $\hat{F}_R$ are fused tokens after cross attention in the two branches. 

After the cross-attention, we reshape and upsample $\hat{F}_D$, $\hat{F}_R$ to the same scale of the current input feature map.
Moreover, we utilize a learned weighted-sum in the input feature maps and fused attention maps for preserving the original features.
Generally, the feature fusion module allows one domain to help guide the direction of another domain's generation with consistent features. The clear edges of RGB images can efficiently help preserve semantic boundaries in the depth generation results.
Meanwhile, we also generate RGB images with a more visual depth hierarchy.


\section{Experiments}

In this section, we first introduce the details of our experiments, including dataset, evaluation metrics, loss functions, and implementation details. Then, we provide the qualitative and quantitative comparisons of our method with state-of-the-art CNN-based models. Finally, we perform the ablation studies to validate the effectiveness of our network structure.

\subsection{Experiments Details}
\begin{figure}[htbp]
    \begin{center}
	    \includegraphics[width=0.4\textwidth]{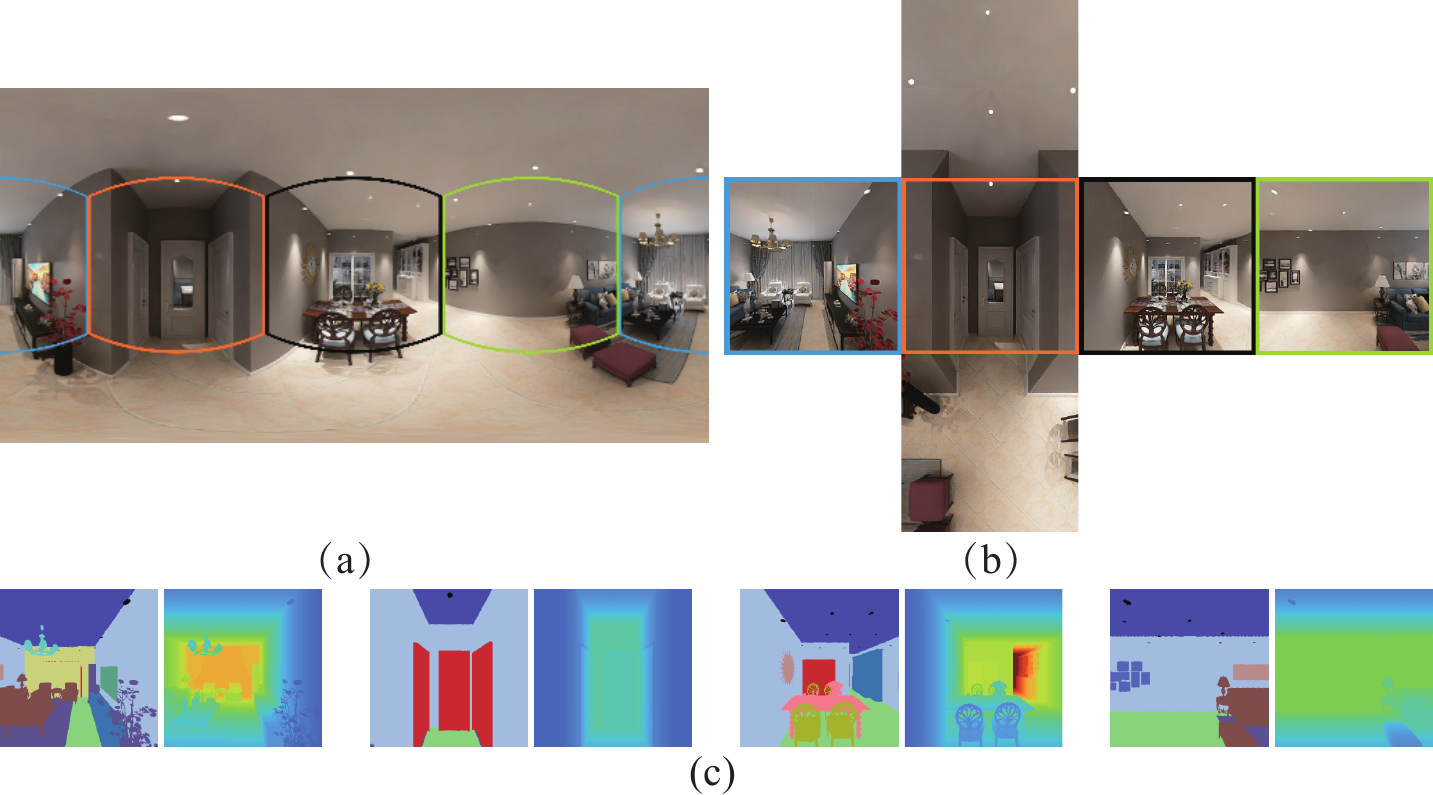}
    \end{center}
    \caption{Explanation of our data preparation process. (a) Panoramic image in Structured3d dataset. (b) Our data in cubemap. (c) Semantic and depth map pairs in the scene, viewed in color. The mapping between (a) and (b) with the same color box represents the transformation from a spherical view to a cubic view.}
	\label{fig:en2cube}
\end{figure}

\paragraph{Dataset.} We conduct our experiments on the Structured3d dataset \cite{ref:structured3d}, which
contains 21835 RGB-D virtual indoor panoramic images with detailed semantic and depth annotations. For our depth generation, we transform the panoramic images from panorama to perspective views as shown in Fig.\ref{fig:en2cube} (a) and (b). Then we choose the images except ceiling and floor in a cubemap with their semantic layouts and depth maps as the compositions of our post-processed dataset (see Fig.\ref{fig:en2cube} (c)). For other two views, they are meaningless for the depth generation task as their depth are consistent in a perspective view. Furthermore, we follow the official scene train-test split in Structured3d and obtain 72996 images for training and 6748 images for testing. 

\begin{figure*}[htbp]
    \begin{center}
	    \includegraphics[width=0.9\textwidth]{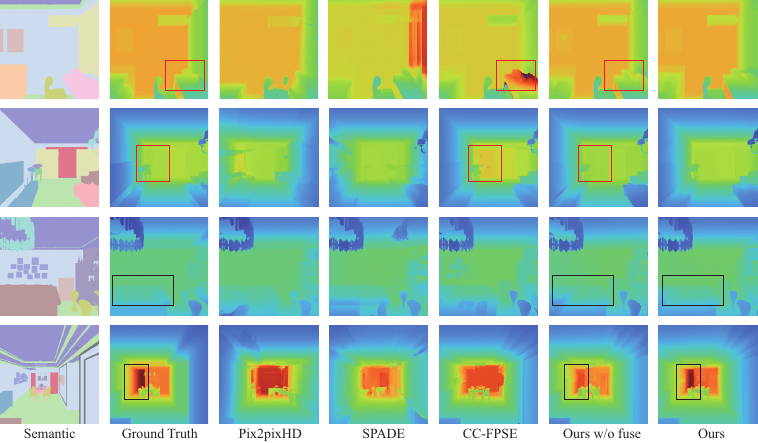}
    \end{center}
    \caption{Visual comparison of depth generation from semantic layouts on the Structured3d dataset with 256$\times$256 resolution. We map the depth value with turbo color mapping, the colors from blue to red mean the process of depth increasing.}
	\label{fig:comp}
\end{figure*}

On the other hand, the depth maps in Structured3d are stored as 16-bit gray-scale images while the data for the generation task is usually in the format of an 8-bit image. In order to accurately describe the depth structure of each view and maintain the numerical accuracy, we normalize the processed cubemap by dividing its maximum depth value and rescale each view to the range of (0,255), and we store the perspective depth map as a 32-bit float number.

\paragraph{Evaluation Metrics.} We adopt six standard evaluation metrics commonly used in depth estimation tasks \cite{ref:error,ref:AdaBins}, including mean absolute error (MAE), root mean square error (RMSE), logarithmic root mean square error (RMSElog), absolute relative error (Abs Rel), square relative error (SqRel) and threshold percentage ($\delta^n$). Note that, we align the generated depth with ground truth by the mean value during evaluation since the ill-posed problem of monocular depth generation. Thus, we concentrate more on the structure information of the depth map. The details of evaluation metrics are provided in the supplementary.




\paragraph{Loss Functions.}
Our DepthGAN uses a generator and two discriminators to generate and discriminate two generating branches respectively.
First, we choose the multi-scale discriminator in \cite{ref:pix2pixHD}
for both depth map and RGB image branches. 
Then, the generator and discriminators of our network are trained alternatively. 
Our generator is optimized with multiple losses used in \cite{ref:SPADE}, including hinge-based adversarial loss, GAN feature matching loss, perceptual loss for RGB branch while structural similarity loss(SSIM) \cite{ref:ssim} for depth branch.
And all discriminators adopt the GAN hinge loss \cite{ref:GeoGAN} for distinguishing real/fake images.
\paragraph{Implementation Details.}
For the generator, we adopt synchronized batch normalization between different GPUs for better estimating the batch statistics. For both discriminators, we apply the Spectral Norm to all the layers. We adopt Adam optimizer with learning rate 0.0001 for the generator and 0.0004 for both discriminators, and we set $\beta_1=0$ and $\beta_2=0.999$. The weight for the perceptual loss in RGB branch is 10 and the weight for the SSIM loss in depth branch is 20. All experiments were conducted on 
8 NVIDIA TITAN RTX 24GB GPUs. More details are provided in the supplementary material.

\subsection{Comparison Experiments}
In this sub-section, we provide the quantitative and visual comparisons on the post-processed Structured3d dataset to prove the effectiveness of our DepthGAN.
We compare with three leading CNN-based conditional image generation models including pix2pixHD \cite{ref:pix2pixHD}, SPADE \cite{ref:SPADE}, and CC-FPSE \cite{ref:CC-FPSE} at the 128$\times$128 and 256$\times$256 resolutions respectively. For a fair comparison, all performances are reported at 100 epochs with the same training strategy as ours.
We also conduct our experiments on both single depth branch and two branches with cross attention fusion.


\paragraph{Qualitative Results.}


As shown in Fig.\ref{fig:comp}, our DepthGAN generates more accurate depth maps with reasonable structure correlation and semantic boundaries. 
Furthermore, the depth hierarchical structure of our results is closer to the ground truth. 
Meanwhile, our transformer architecture generates depth maps with more reasonable structural correlations (see the red boxes in the first two rows). 
Pix2pixHD and SPADE always generate additional objects with incorrect depth relationships, and the depth maps may be over-smoothed.
Moreover, the table and curtain in the first and second row are generated incorrectly for CC-FPSE. Our results have boundaries that are neither blurred nor over-sharpened as shown in the last two rows of Fig.\ref{fig:comp}.


\begin{table*}
\centering
\begin{tabular}{c|c|ccccccccc}
\toprule
Resolution & Method  & MAE$\downarrow$ &  AbsRel$\downarrow$ & SqRel$\downarrow$ & RMSE$\downarrow$ & RMSElog$\downarrow$ & log10$\downarrow$ & $\delta_1\uparrow$ & $\delta_2\uparrow$ & $\delta_3\uparrow$\\
\midrule
\multirow{5}*{128} & Pix2pixHD & 0.1699  & 0.1561 & 0.1397 & 0.2352 & 0.0901 & 0.0610 & 0.8307 & 0.9203 & 0.9535 \\
~ & SPADE & 0.1508  & 0.1340 & 0.1075 & 0.1945 & 0.0698 & 0.0492 & 0.8667 & 0.9464 & 0.9740  \\
~& CC-FPSE & 0.0963  & 0.0851 & 0.0365 & 0.1368 & 0.0527 & 0.0368 & 0.9158 & 0.9739 & 0.9874   \\
~& Ours w/o fuse & 0.0939 & 0.0795 & 0.0348 & 0.1357 & 0.0490 & 0.0335 & 0.9196 & 0.9734 &  0.9888 \\
~ & Ours & \textbf{0.0873}  & \textbf{0.0764} & \textbf{0.0336} & \textbf{0.1278} & \textbf{0.0467} & \textbf{0.0328} & \textbf{0.9205} & \textbf{0.9749} & \textbf{0.9903}    \\ 
\midrule
\multirow{5}*{256} & Pix2pixHD & 0.1701  & 0.1553 & 0.1207 & 0.2349 & 0.0524 & 0.0608 & 0.8332 & 0.9257 & 0.9589 \\
~ & SPADE & 0.1749 & 0.1497 & 0.1261 & 0.2405 & 0.0610 & 0.0507 & 0.8381 & 0.9203 & 0.9535 \\
~ & CC-FPSE & 0.1111 & 0.1018 & 0.0510 & 0.1594  & 0.0630 & 0.0418 & 0.9023 & 0.9672 & 0.9841  \\
~ & Ours w/o fuse & 0.1013 & 0.0905 & 0.0386 & 0.1426 & 0.0516 & 0.0374 & 0.9149 & 0.9763 & 0.9870 \\
~ & Ours & \textbf{0.0972} & \textbf{0.0855} & \textbf{0.0371} & \textbf{0.1387} & \textbf{0.0472} & \textbf{0.0343} & \textbf{0.9199} & \textbf{0.9773} & \textbf{0.9905} \\ 
\bottomrule
\end{tabular}
\caption{Quantitative comparison results by the proposed and previous methods on our post-processed Structured3d dataset at different resolutions. For the last three metrics, higher is better.}
\label{tab:comparation}
\end{table*}

\begin{table*}
\centering
\begin{tabular}{c|cccccccccc}
\toprule
Method  & MAE$\downarrow$ &  AbsRel$\downarrow$ & SqRel$\downarrow$ & RMSE$\downarrow$ & RMSElog$\downarrow$ & log10$\downarrow$ & $\delta_1\uparrow$ & $\delta_2\uparrow$ & $\delta_3\uparrow$\\ 
\midrule
w/o patch & 0.1107 & 0.0973 & 0.0530 & 0.1561 & 0.0549 & 0.0385 & 0.9047 & 0.9680 & 0.9855 \\
Fixed patch & 0.1102 & 0.1002 & 0.0585 & 0.1552 & 0.0552 & 0.0387 & 0.9052 & 0.9659 & 0.9839 \\
w/o fuse & 0.1036 & 0.0919 & 0.0437 & 0.1480 & 0.0538 & 0.0372 & 0.9119 & 0.9726 & 0.9879 \\
w/o ssim & 0.1048 & 0.0907 & 0.0447 & 0.1522 & 0.0541 & 0.0370 & 0.9099 & 0.9731 & 0.9887 \\
Increasing patch(Ours) & \textbf{0.0991} & \textbf{0.0860} & \textbf{0.0397} & \textbf{0.1402} & \textbf{0.0481} & \textbf{0.0354} & \textbf{0.9186} & \textbf{0.9778} & \textbf{0.9899} \\
\bottomrule
\end{tabular}
\caption{Ablation studies on our post-processed Structured3d dataset at resolution 256$\times$256.}
\label{tab:ablation}
\end{table*}

\paragraph{Quantitative Results.}
As shown in Tab.\ref{tab:comparation}, 
We have improved performance by more than 10\% over the state-of-the-art methods under our proposed depth generation settings, especially for the second-order metrics such as \text{SqRel} and \text{RMSE}.  This indicates our results have fewer extreme depth values due to the perception of global information by the transformer-based architecture.
On the other hand, we also achieve higher $\delta_n$ and lower first-order metrics.
Meanwhile, a large margin exists between the former two methods and other methods since there are no enough constraints on global features or semantic consistency.
In particular, our method achieves more precedence at resolution 256$\times$256 than 128$\times$128 as the superiority of the transformer architecture in larger resolutions.

\subsection{Ablation Studies}
To validate the effectiveness of our DepthGAN, we conduct
ablation studies at 256$\times$256 resolution as shown in Tab.\ref{tab:ablation}. All performances are reported using training results of 60 epochs.

\paragraph{Cross Attention Fusion.}
We verify the effect of our attention fusion strategy by removing the color image generation branch and the CAF module, respectively. As shown in Fig. \ref{fig:comp}, compared to results without fusion, our results marked by the black boxes in the third and fourth rows reveal that the appearance of RGB images can efficiently help preserving edges in the depth maps especially for objects with complicated semantics such as each individual pillow on sofa and complex wall structure.

\paragraph{SSIM Loss Function.}
We replace the loss of depth branch with perceptual loss in \cite{ref:SPADE} instead of the SSIM loss, which leads to a slight decrease of model performance.
The perceptual loss focuses more on the global features of the image, while the SSIM loss cares about the difference between two images more from the pixel level. 
Thus, using SSIM loss contributes to better results for the generation of depth maps.


\paragraph{Patch Size in Generator.}
We select three different strategies of patch size in our generator blocks: (1) Without patch embedding; (2) Fixed patch size at each stage such as 4; and (3) The patch size increases with the input feature resolution. We empirically set the patch size to be ${1}/{32}$ of input resolution for the 128$\times$ 128 task and ${1}/{64}$ of input resolution for the 256$\times$ 256 task while the minimum value is 1.  

As shown in Tab.\ref{tab:ablation}, the relative errors(AbsRel,SqRel) and $\delta_{n}$ of strategy(1) are better than strategy (2), which indicates that the pixel accuracy of the former is greater than that of the latter and the probability of generating extreme depth values is smaller.  
Finally, our increasing strategy generates the best results
in numerical metrics. Since our patch size increasing strategy embeds the feature map at different stages to the same size with the enlarging window size, which mimics the receptive field in CNNs, we can efficiently generate depth details of objects at different scales.


\section{Conclusion}
In this work, we first propose the DepthGAN to efficiently generate the depth maps for indoor scenes from the input of semantic layouts, which provides an effective solution for complex 3D scenes generation. Meanwhile, our depth generation results outperform the SOTA CNN-based image generation approaches in terms of structural correlations and depth edge preservation.

In addition, the results for fine components of objects are still relatively coarse. In our future work, we will further improve the depth generation for the fineness of objects.
Furthermore, we will also consider simplifying the input to scene graphs or texts and applying our method to outdoor scenes.



\bibliographystyle{named}
\bibliography{ijcai}

\end{document}